# Investigating LLM's Problem Solving Capability - a Study on Statics Questions

**Tanner Culleton[1,*], Hung-Fu Chang[1]**

[1]*R. B, Annis School of Engineering, University of Indianapolis, Indianapolis, IN, 46227, USA*

*Corresponding's Email:* tculleto@uindy.edu

***Abstract*— Large Language Models (LLMs) have rapidly influenced many aspects of society, particularly education, due to their demonstrated ability to complete assignments and examinations across a wide range of subjects. Although prior studies have examined the educational impact of LLMs, much of the existing work relies on public or open problem datasets and lacks topic-specific analysis. In engineering education, especially within mechanical engineering, systematic investigations of LLM performance on specific problem types remain limited. Instead of using traditional methods that directly ask textbook questions to an LLM tool, our study adopts a model distillation process to evaluate LLM capabilities in solving statics problems. By distilling ChatGPT, we extracted 25 text-only statics questions and further constructed two additional datasets by adding diagrams and modifying their numerical values. Experimental results show that while LLMs perform well on text-only statics problems, their accuracy decreases when diagrams are introduced and the problems require multi-step reasoning. Further analysis suggests that this performance drop is not primarily caused by limitations in image recognition, but rather by difficulties in multi-step reasoning and in consistently applying extracted visual information across successive solution stages.**

## I. INTRODUCTION

Large Language Models (LLMs) have rapidly influenced many aspects of society, with particularly significant implications for education [1]. One primary reason for the growing attention is the demonstrated capability of LLMs to complete assignments and examinations in various subjects. LLMs often achieve a competitive level of correct task completion compared to human learners. This capability has raised important questions regarding assessment validity, academic integrity, and the role of artificial intelligence in supporting or hindering student learning [2].

While numerous studies have explored the educational impact of LLMs, much of the existing work remains public or open problem sets and lacks textbook or topic-specific analysis. In particular, systematic and in-depth investigations within engineering education, especially those focused on specific problem types in mechanical engineering, remain limited [3]. In addition, given the rapid pace at which LLM capabilities continue to evolve, it is essential to evaluate recently developed models to ensure that educational research reflects their current performance characteristics [4, 5].

Unlike traditional approaches that assess LLMs primarily through direct performance assessments, which is asking LLMs to answer the questions on academic tests or professional exams, our approach adopts the perspective of model distillation [6] in the evaluation process for analyzing LLM performance in an educational context. Model distillation is a technique that aims to extract training instances from a well-performed model and then use the extracted data to train another language model. This teacher-and-student model relationship helps to bring the training model, often smaller size, to a similar level of performance.

To evaluate LLM's capability in answering mechanical engineering questions. we start from asking ChatGPT by posing a diverse set of textbook-based statics problem to have basic understanding of ChatGPT's problem solving capability. We then apply a distillation-inspired process to identify questions that the model consistently answers correctly to create a question dataset. With manually altering each question with drawing and new numerical values, we can further investigate how LLM reasoning works. By identifying both the capabilities and limitations of the LLMs and understanding which types of problems LLMs can readily solve enables educators to design assignments and assessments that better promote conceptual understanding and higher-order thinking, while mitigating inappropriate reliance on AI tools [7]. Ultimately, this work aims to inform educators and curriculum designers on how to adapt instructional strategies and assessment practices in response to the evolving capabilities of LLMs.

## II. RELATED RESEARCH

### A. *LLM Problem Solving Ability in Math*

Early GPT-3 models demonstrated limited ability on complex mathematical reasoning tasks; however, substantial improvements emerged in 2022 with specialized and larger-scale language models. Google's Minerva, a transformer further trained on scientific and mathematical data, achieved state-of-the-art results on quantitative benchmarks, solving approximately 50% of problems in the MATH competition dataset, compared to roughly 7% for prior models [12]. Minerva also reached 78.5% accuracy on GSM8K and strong performance on STEM subsets of MMLU, with gains attributed largely to chain-of-thought prompting and sampling-based answer selection strategies. General-purpose models such as GPT-4 have also shown strong quantitative performance on standardized exams, including AP Calculus and SAT Math, substantially outperforming GPT-3.5 [13]. Nevertheless, performance remains inconsistent on high-difficulty or competition-style problems (e.g., AMC-level questions), where GPT-4 typically ranks well below expert human performance. Error analyses indicate that failures arise from both arithmetic mistakes and breakdowns in long multi-step reasoning, exacerbated by the absence of external verification tools or symbolic solvers [12, 13]. Overall, while LLMs approach near-expert performance on structured math exams, they remain unreliable for problems requiring rigorous step-by-step correctness.

### *B. LLM in Answering Law Exam Questions*

Initial evaluations of GPT-3.5 on law school examinations revealed mixed performance. In blind grading experiments conducted at the University of Minnesota Law School, ChatGPT earned an average grade of approximately C+, passing but underperforming relative to the student mean [14]. Qualitative analysis showed that the model produced coherent and well-structured essays and accurately stated many legal rules; however, it frequently failed to identify key issues or apply doctrine with sufficient depth. Performance on multiple-choice questions exceeded random guessing but was weaker on numerically intensive legal problems, such as damages calculations. Subsequent evaluations showed dramatic improvements with GPT-4. OpenAI reported that GPT-4 achieved a score corresponding to approximately the 90th percentile on a simulated Uniform Bar Exam and performed strongly on the LSAT [13]. However, legal scholars caution that such results should be interpreted carefully. The bar exam is closed-book for human test-takers but effectively open-book for LLMs trained on large legal corpora, raising concerns about memorization and data leakage. Moreover, standardized exams primarily test general legal knowledge and issue spotting, whereas real-world legal practice requires strategic judgment, client interaction, and ethical decision-making that remain beyond current models' capabilities [15].

### *C. LLM in Answering Physical Engineering Questions*

Studies in mechanical and civil engineering suggest that LLMs can perform well on structured, textbook-style problems, but exhibit limitations on open-ended or visually complex tasks. A 2024 study evaluating GPT-4 on an undergraduate engineering statics course found that, with optimized prompting and explicit reasoning guidance, the model achieved an average score of 82% on a final exam, exceeding the class average of 75% [16]. These results demonstrate that LLMs can effectively apply formulas and analytical procedures under controlled conditions. Despite this success, significant weaknesses remain. GPT-4 struggled with problems requiring qualitative physical reasoning, such as identifying tension and compression in truss members, and often produced plausible but physically incorrect solutions. Performance also degraded substantially when problems relied on diagrams rather than textual descriptions, with accuracy dropping to approximately 50% in some cases. These findings indicate that current text-only LLMs lack robust visual-spatial reasoning and true physical understanding, limiting their reliability in safety-critical engineering contexts.

### *D. LLM in Programming*

Programming and algorithmic problem-solving remain challenging domains for LLMs. In 2022, DeepMind introduced AlphaCode, which combined large-scale language models with extensive sampling and filtering to solve competitive programming problems. AlphaCode achieved approximately median human performance on Codeforces competitions, ranking in the top 54% of participants [19]. This demonstrated that AI systems could reach average human-level performance when augmented with search and selection mechanisms. By contrast, general-purpose models such as GPT-4 perform well on simple coding tasks but struggle with more complex algorithmic problems. According to the GPT-4 Technical Report [13], the model solved most easy programming problems but fewer medium-level problems and only a small fraction of hard ones. Without execution feedback or iterative debugging, GPT-4 often produces code that is syntactically correct but logically flawed. As a result, while LLMs are effective as coding assistants, they remain unreliable for solving complex programming challenges independently.

## III. METHODOLOGY

### *A. Overview of Our Study*

Our overview process of conducting the investigations can be seen in Figure 1. It starts with initial analysis on the textbook questions that we copy and paste to ChatGPT to see how good it can solve the problem. However, we learned that ChatGPT cannot answer the textbook questions very well so that we switch to see if ChatGPT can answer its own generated questions well. With the understanding of how ChatGPT answers its own generated questions. We build up three datasets where questions are all two-dimensional statics problems because ChatGPT cannot provide enough three-dimensional questions and cannot solve any of their questions when they are converted into diagram. After constructing all the datasets, we do our evaluation on ChatGPT and Gemini.

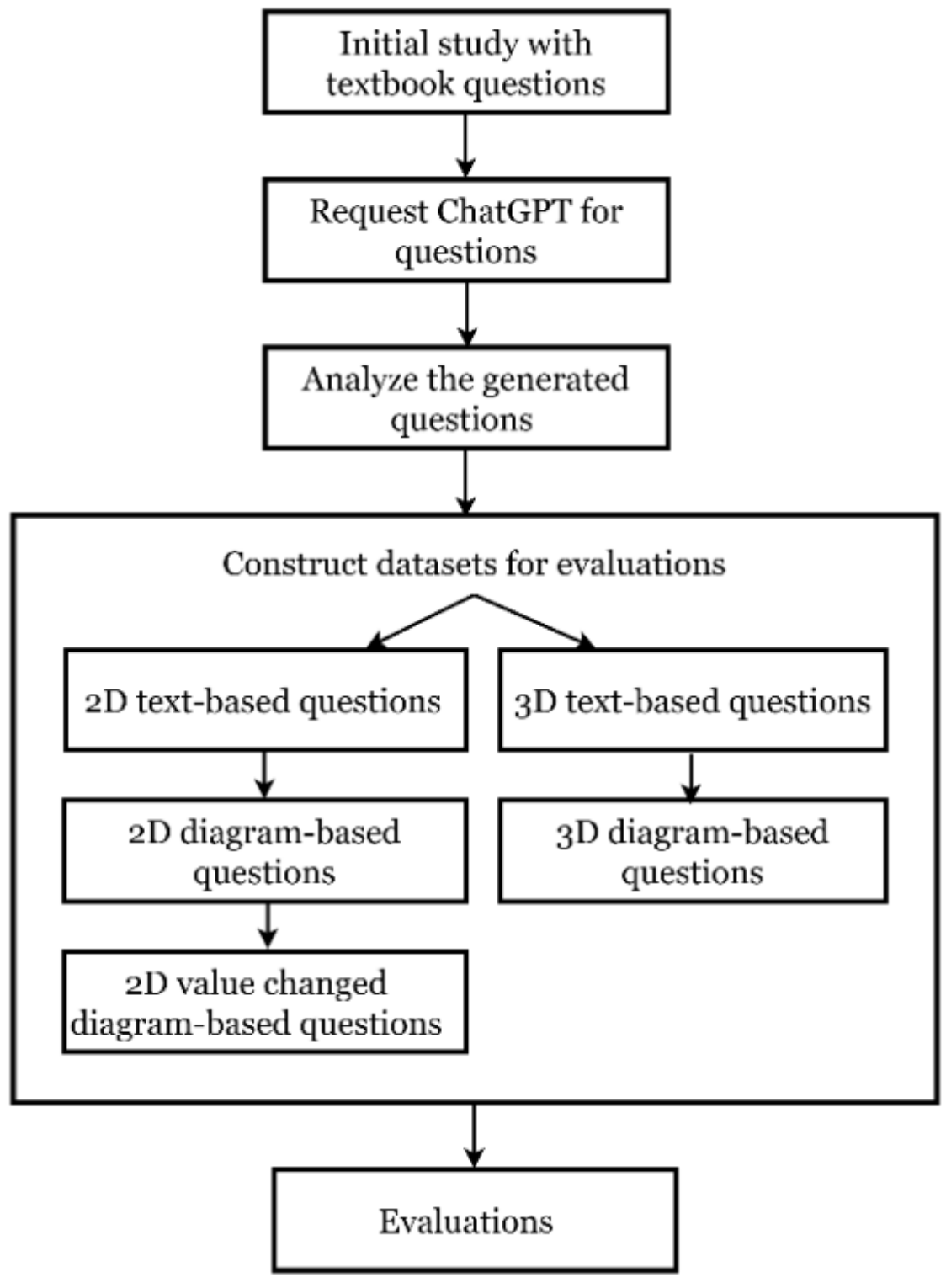

*Figure 1: Overall flow of our study*

## B. Lesson Learned From Direct Testing

To have a basic understanding of LLM's reasoning ability, we start with directly ask ChatGPT to statics questions that extract from an online textbook from "McGraw Hill Connect" [20]. The first observation from initial testing of ChatGPT's performance on statics problems is that three-dimensional problem seems to be as its primary limitation. We suspect the difficulty arises from mainly problem comprehension or diagram interpretation. It might be inherently difficult for ChatGPT to comprehend what the question is about. Another cause is incorrect assignment of vector components between coordinate axes done by ChatGPT. ChatGPT would mix up the vectors in the equations by placing the information that needed to be in the "x" direction to be in the "z" direction and the information for the "y" direction to be in the "x" direction. The false vector assignment leads to our assumption of limited ability in image interpretation for LLMs. Unlike humans, who naturally perceive depth, ChatGPT processes diagrams as two-dimensional representations and must infer three-dimensional relationships, increasing the likelihood of misinterpretation.

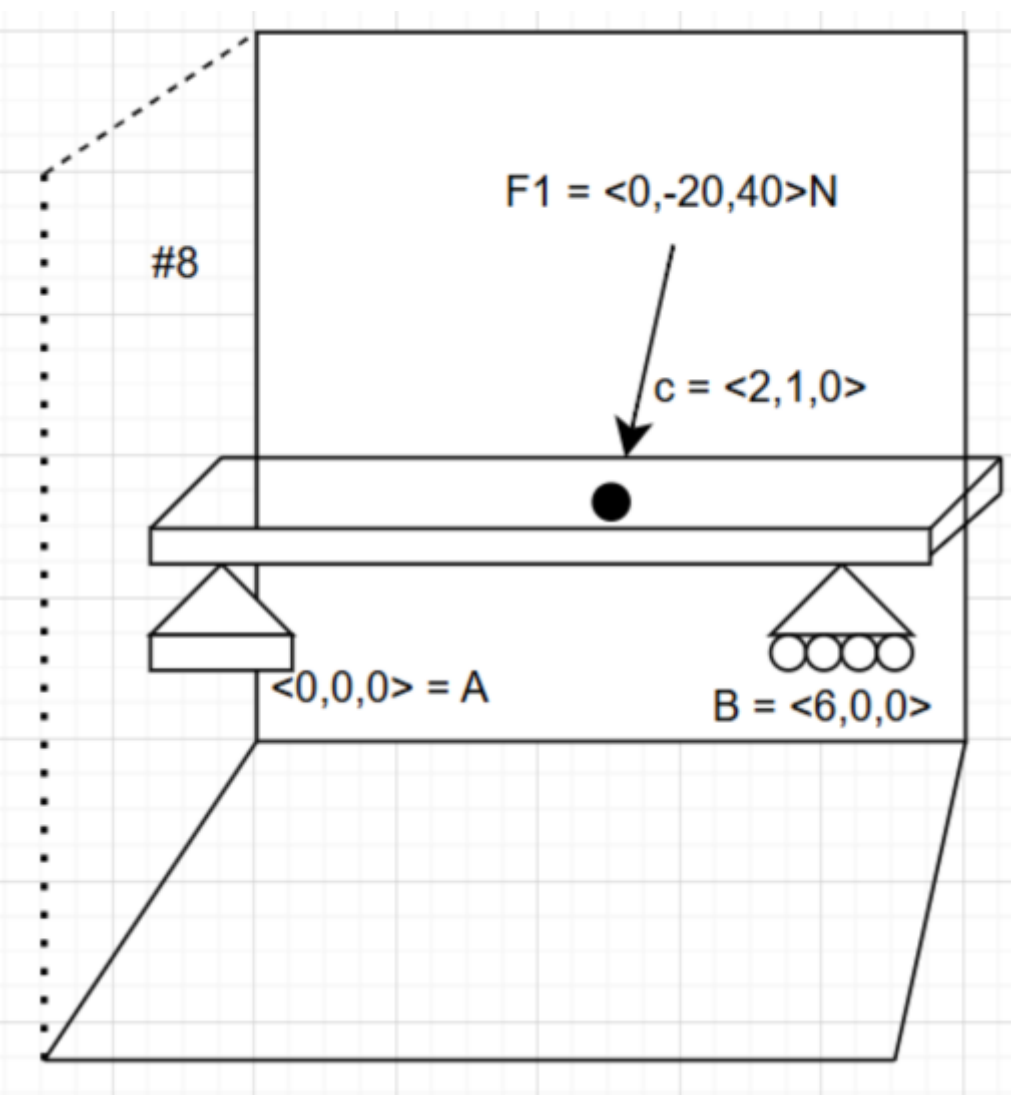

*Figure 2: An example of three-dimensional questions that GPT answer incorrectly.*

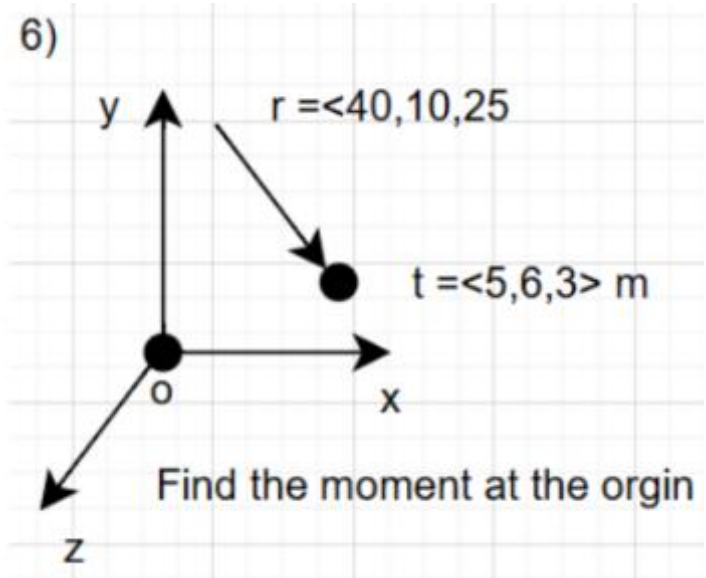

*Figure 3: An example of three-dimensional questions that GPT answer correctly.*

From this early results, we learn that ChatGPT is better at understanding three dimensional problems if the drawing is done clearly, coordinates are explicitly expressed in vector form, and key aspects of the problem are described verbally. Additionally, when the three-dimensional problem needs in-depth reasoning, ChatGPT cannot answer it correctly. Figures 2 and 3 show a incorrectly-solved and a correctly-solved example, respectively. We also observe that ChatGPT produced incorrect numerical results, but it still generally followed the correct problem-solving procedures; the errors were primarily in the calculations rather than the methodology. After determining that ChatGPT is not consistently reliable when solving three-dimensional statics textbook problems, we hypothesized that this limitation may be due to insufficient representation of three-dimensional statics in its training data. To investigate this, we first probed the types of statics questions that may be embedded within the model by repeatedly prompting ChatGPT to generate sets of 50 statics problems. This approach allowed us to examine the kinds of questions ChatGPT is capable of producing and, by extension, the types of statics problems likely reflected in its training distribution.

After generating problems across multiple prompts, it became clear that relatively few of the questions were complex. Even without assessing the accuracy of ChatGPT's solutions, analysis of the generated problems showed that most were short, straightforward, and often limited to a single sentence (see Figure 4). Most of the questions were two-dimensional in nature. Additionally, ChatGPT frequently reused the same physical objects across different categories or created questions with different values within the same type. This behavior indicates its limited diversity in both problem structure and context.

### Rigid Bodies – Beams and Rods

16. A uniform beam of length 4 m and weight 200 N rests on two supports at its ends. Find the reaction at each support.
17. A uniform beam of length 6 m and weight 300 N is supported at its ends. A 500 N load is placed 2 m from one end. Find the reactions at the supports.
18. A 5 m long beam of weight 400 N is hinged at one end and supported by a vertical string at the other end. Find the tension in the string.
19. A uniform rod of length 3 m and weight 150 N is hinged at one end and held horizontal by a cable attached at the free end making 30° with the rod. Find the tension in the cable.
20. A beam of length 8 m is supported at its ends. A 600 N load acts at 3 m from the left end. Find the reactions.

### Moments and Torque

21. A 50 N force acts at a perpendicular distance of 0.4 m from a pivot. Find the moment of the force.
22. A uniform rod of length 2 m and weight 100 N is pivoted at one end. Find the moment of the weight about the pivot.
23. Two forces of 20 N act in opposite directions at a distance of 0.5 m apart. Find the moment of the couple.
24. A beam is in equilibrium under a clockwise moment of 120 N·m. Find the anticlockwise moment required for balance.
25. A force of 30 N acts at 60° to a rod at a point 0.8 m from the pivot. Find the moment about the pivot.

*Figure 4: Example results given by asking GPT for 50 statics questions to solve*

Since large language models may perform poorly on certain problems due to gaps or biases in their training data, this research shifted away from directly testing an LLM on externally provided questions. Instead, the focus moved toward understanding what types of statics problems are represented within the model itself and whether the LLM can successfully solve problems that are likely embedded in its training distribution. This perspective allows us to better evaluate whether an LLM reasons through statics problems in a manner similar to human problem-solving. To accomplish this, we adopted an approach inspired by model distillation by prompting ChatGPT to generate a large and diverse set of statics problems. These generated questions were then used to assess whether GPT could demonstrate consistent and effective problem-solving performance on problem types aligned with its internal knowledge.

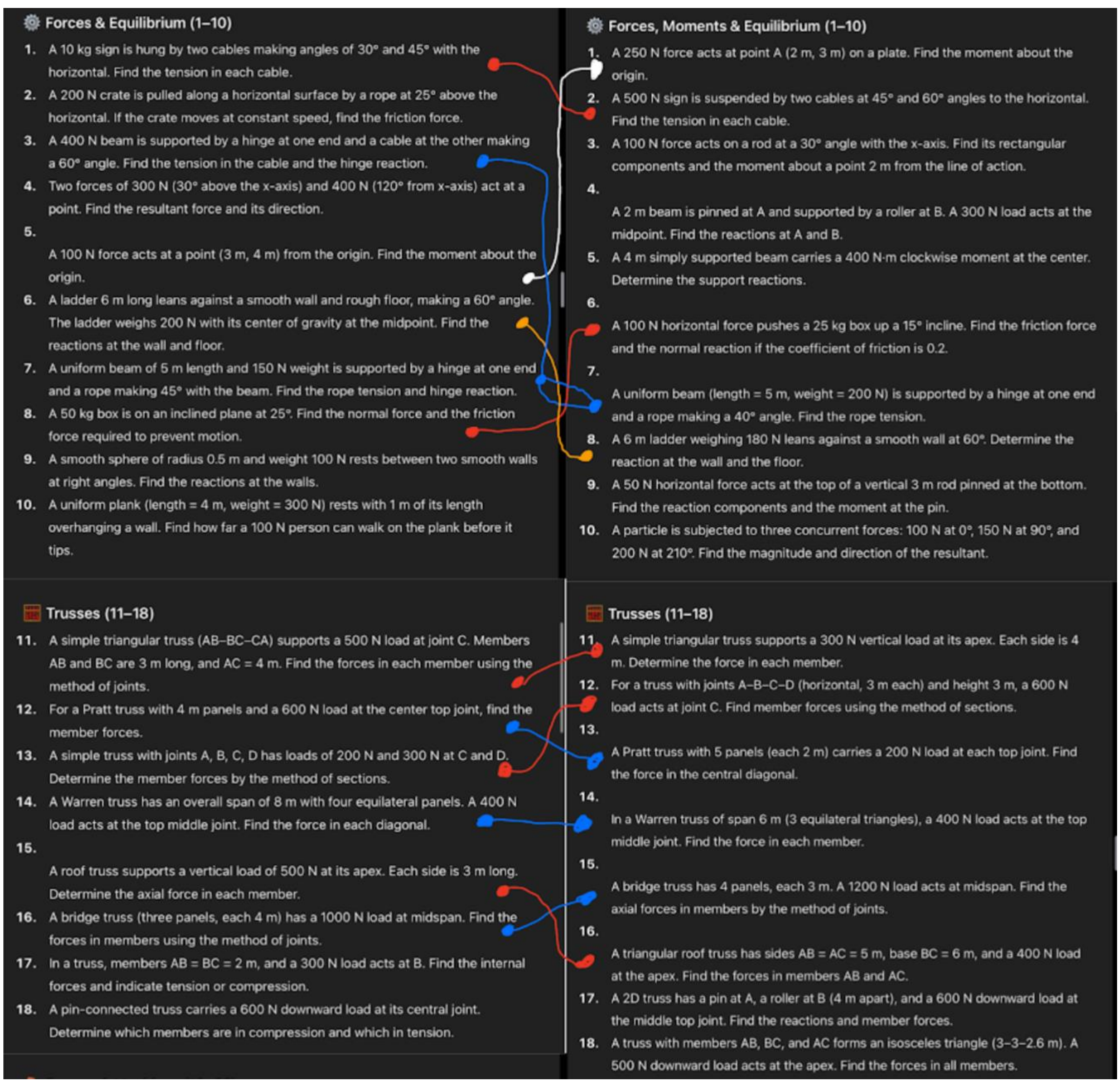

*Figure 5: A portion of two sets of 25 questions with comparing answers. In the figure, different color connected lines mean that two questions are related after multiple times of requests.*

*C. Eclicit Question for Model Evaluation Creation from Model Distillation*

By knowing that GPT tends to generate repetitive questions after a certain point from our initial investigation. Figure 5 shows the same questions with value changes or even the same questions we identify through multiple times of generation. Therefore, the next step is to determine the number of questions we can extract for evaluation. Two criteria are used to decide the number of questions. The data needed to be large enough to meaningfully assess the model's performance while remaining feasible within time constraints. To determine this number, a structured plan was developed by using a distillation-inspired approach. During the question extraction, ChatGPT was repeatedly prompted to provide "new" statics questions, and each new set was compared with previously generated questions. This process continued until the questions became clearly repetitive. Based on this procedure, a final set of 25 questions was selected for evaluation. These questions are all two-dimensional questions with only textual description. The types of questions that we identified and also categorized by difficulty (see Table 1). Eight problems were classified as easy, involving straightforward equilibrium conditions, single concepts, and minimal algebra. Seven problems were considered medium difficulty, requiring multiple equilibrium equations, and careful selection of moment points. The remaining ten problems were classified as difficult, as they combined multiple statics concepts and multi-step reasoning, and required strategic problem setup.

*Table 1: Level of Difficulty for Each Question Types*

| Question Type | Description | Level of Difficulty | Number of Questions |
|---|---|---|---|
| Conceptual | Questions that test understanding of definitions, physical meaning, or qualitative reasoning without heavy calculations. | Easy | 8 (Question: 1, 2, 3, 4, 11,12,13, 14) |
| Direct Calculation | Problems that require applying known formulas and performing straightforward calculations with given values. | Medium | 7 (Question: 15, 16, 17, 18, 20, 21, 24) |
| Multi-Step Application | Problems that involve multiple steps, combining concepts, or interpreting diagrams or setups before solving. | Hard | 10 (Question: 5, 6, 7, 8, 9 ,10, 19, 22, 23, 25) |

*Table 2: Evaluation Dataset*

| Name | Creation Method | Evaluation Goal |
|---|---|---|
| Text-only Dataset | Originally generated from ChatGPT | To test if the LLM can solve the problem described in text only. |
| Diagram-based Dataset | Converting the text-only questions with diagrams. For some questions, text description is provided to indicate what values need to be calculated. | To test if the LLM can understand diagram when the same question is presented. |
| Value-changed Diagram-based Dataset | Replace the diagram-based questions with new numerical values | To test if the LLM can correctly calculate the question after numerical values are replaced with new ones. |

Since 25 ChatGPT generated statics problems only contain textual description without providing diagrams, we also want to evaluate its ability to solve the same problems when drawn figures were presented. We name these generated questions as text-only dataset, meaning these are originally generated by ChatGPT (see Table 2). To create a new dataset with drawing, all textual descriptions in the question were manually converted into diagrams. These new set of question will be reintroduced to an LLM to assess whether the model could recognize and correctly interpret problems it had previously generated. The diagrams were created using the *Notability* app and were drawn as cleanly and simply as possible to ensure the LLM could accurately interpret the visual information (see Figure 6). When necessary, brief written descriptions were added to clarify the intent of each problem. This collection of diagram-based problems is referred to as the diagram-based dataset (see Table 2).

To gain further insight into ChatGPT's ability to interpret images of statics problems, the questions in diagram-based dataset were modified again. Specifically, the diagrams were edited by changing the numerical values of forces, moments, and distances. This modification was intended to determine whether ChatGPT was recalling solutions from prior exposure or genuinely understanding and solving the problems based on the visual information provided (see Table 2). These modified questions are designed to be submitted to ChatGPT using the same prompt as the original diagram-based problem. Figure 6 shows an example of how a text-only question is converted to a diagram-based problem and Figure 7 demonstrates how the same diagram-based questions is changed to a value-changed diagram-based problem by replacing the numerical values.

**14 - Beam with triangular distributed load**

Problem: Beam AB length 6.00 m simply supported. Triangular distributed load with intensity zero at A and $w_o$ = 12 kN/m at B (linear along beam). Find reaction $A_y$ and $B_y$.

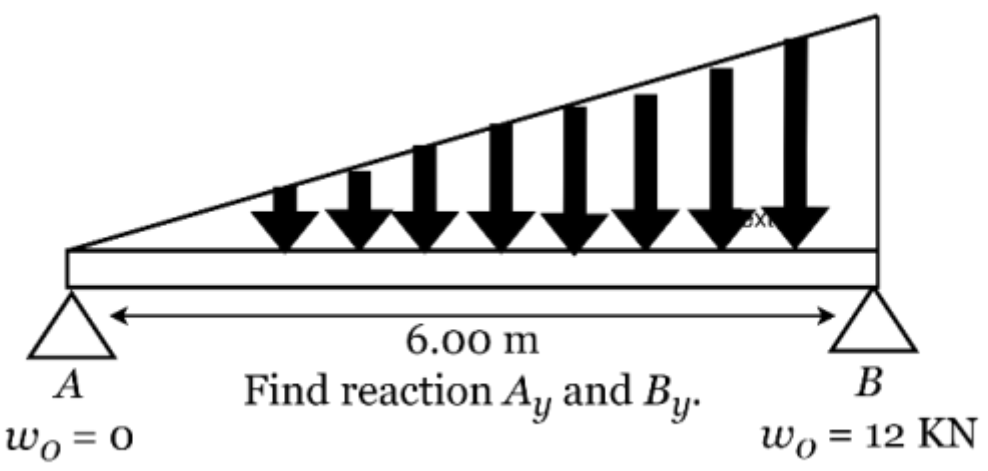

*Figure 6: An example of problem conversion from text-only to diagram-based*

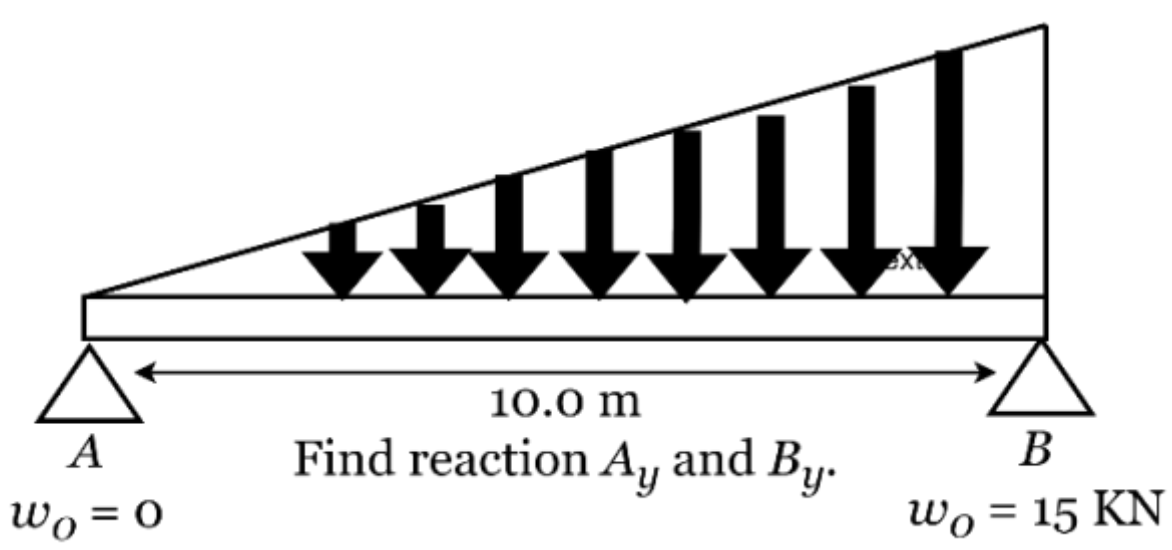

*Figure 7: An example of value-changed diagram-based questions converted from question in Figure 6.*

### D. Measurement

The measurement used in this study was model accuracy, defined as the number of questions correctly solved by the LLM tool $N_c$ out of total number of questions $N_T$. Eq. (1) shows how accuracy is calculated. Given that the number of the total questions in our dataset is 25, Eq. (1) can be converted to Eq. (2)

$$Accuracy\ A = N_c/N_T \tag{1}$$

$$Accuracy\ A = N_c/25 \tag{2}$$

## IV. Results and discussions

### A. Result Analysis

Based on the testing results for the text-only dataset, ChatGPT-5 achieved 100% accuracy, whereas Gemini achieved 60% accuracy. This outcome was expected because the questions in the text-only dataset were originally distilled from ChatGPT-5. Notably, Gemini was unable to correctly solve any of the multi-step application problems, resulting in 0% accuracy in this category. We suspect that both ChatGPT-5 and Gemini were trained on datasets containing basic and straightforward questions, but they may differ in their preparation for multi-step application-based problems. When evaluated on the diagram-based dataset, ChatGPT-5 correctly solved 17 out of 25 problems, resulting in an overall accuracy of 68%. More specifically, ChatGPT achieved 100% accuracy on conceptual questions, 71.43% accuracy on direct-calculation questions, and 40% accuracy on multi-step or application-based questions. For the value-changed diagram-based dataset, ChatGPT-5 achieved an overall accuracy of 60%. Performance by question type included 75% accuracy on conceptual questions, 42.8% accuracy on direct-calculation questions, and 60% accuracy on multi-step application questions.

*Table 3: Correctness Rate*

| | ChatGPT 5 | | | Gemini |
|---|---|---|---|---|
| | **T Dataset** | **D Dataset** | **VD Dataset** | **T Dataset** |
| Total | 100% | 68% | 60% | 60% |
| Conceptual | 100% | 100% | 75% | 100% |
| Direct Calculation | 100% | 71.43% | 42.8% | 100% |
| Multi-Step Application | 100% | 40% | 60% | 0% |

Based on the differences between the diagram-based and value-changed diagram-based datasets, the results indicate that changing numerical values did not significantly affect ChatGPT's ability to interpret the diagrams. The generated solutions typically included a clear sequence of steps illustrating how the result was obtained. Across all test results from both ChatGPT and Gemini, the models consistently followed appropriate solution procedures; however, the final numerical answers were often incorrect. These findings suggest that while LLM tools perform well on text-only statics problems, their accuracy decreases when solving image-based, multi-step statics problems. Moreover, this trend indicates that the observed performance drop is not primarily due to poor image recognition. Instead, it suggests that LLMs struggle with multi-step reasoning and with correctly applying extracted visual information across successive stages of the solution process.

### *B. Limitations*

The use of the distillation method may introduce bias, as the questions used to evaluate ChatGPT were generated by ChatGPT itself. This potential bias, however, is acknowledged. The purpose of this approach is not to provide a fully independent benchmark, but rather to evaluate the model's performance on questions that may reflect knowledge, patterns, or question structures acquired from datasets used during model training. This process is analogous to providing students with potential examination questions in advance and subsequently assessing whether they have effectively learned the relevant material. This interpretation may also help explain why Gemini produced comparatively poor results, as Gemini may not have been exposed to similar question patterns during its training.

We also acknowledge that the dataset size may be limited, as 25 questions may not be statistically sufficient to support broad generalizations. This limitation is difficult to avoid when using a distillation-based approach. The inclusion of repeated questions in the dataset was not considered appropriate. In addition, the difficulty level of the questions was determined through the review of two authors, which may introduce an additional source of subjectivity and potential bias.

## V. CONCLUSION

Evaluating the reasoning abilities of large language models (LLMs) in solving academic problems and answering exam questions is an important topic in education. However, in mechanical engineering, limited research has examined how LLMs approach problem solving. To address this gap, we adopt a method inspired by model distillation to construct three datasets by extracting and transforming questions from ChatGPT. Our results show a clear decline in accuracy when moving from pure text-based questions to converted questions that include diagrams and modified numerical values. Although overall accuracy decreases, comparisons across question types in both the diagram-based and value-change diagram-based datasets indicate that LLMs particularly struggle with multi-step reasoning tasks. The results show that LLMs still lack of sufficient reasoning abilities in deal with statics problems.

## VI. ACKNOWLEDGMENT

We thank for the supports from Presidential Scholar Program of the University of Indianapolis.

## VII. REFERENCE

[1] E. Kasneci, K. Sessler, K.-U. Kühnberger, et al., "ChatGPT for good? On opportunities and challenges of large language models for education," Learning and Individual Differences, vol. 103, Art. no. 102274, 2023.

[2] D. Cotton, P. Cotton, and J. Shipway, "ChatGPT: Implications for assessment and academic integrity," Assessment & Evaluation in Higher Education, 2023.

[3] H. Khosravi, S. Sadiq, et al., "Artificial intelligence in engineering education: A systematic review," IEEE Transactions on Education, 2023.

[4] P. Liang, R. Bommasani, et al., "Holistic evaluation of language models," ACM Transactions on Intelligent Systems and Technology, 2022.

[5] OpenAI, "GPT-4 technical report," OpenAI, Tech. Rep., 2023.

[6] G. Hinton, O. Vinyals, and J. Dean, "Distilling the knowledge in a neural network," arXiv preprint arXiv:1503.02531, 2015.

[7] M. Bearman, R. Ajjawi, et al., "Assessment in the age of artificial intelligence," Assessment & Evaluation in Higher Education, 2023.

[8] V. Sanh, L. Debut, J. Chaumond, and T. Wolf, "DistilBERT, a distilled version of BERT: Smaller, faster, cheaper and lighter," arXiv preprint arXiv:1910.01108, 2019.

[9] T. Taori et al., "Alpaca: A strong, replicable instruction-following model," Stanford CRFM, 2023. [Online]. Available: https://crfm.stanford.edu

[10] Y. Li et al., "Distilling step-by-step reasoning from large language models," arXiv preprint arXiv:2305.02301, 2023.

[11] S. Tang, Y. Liu, and C. Zhang, "Model compression and distillation for edge intelligence: A survey," IEEE Communications Surveys & Tutorials, vol. 25, no. 1, pp. 1–26, 2023.

[12] K. Lewkowycz et al., "Solving Quantitative Reasoning Problems with Language Models," arXiv preprint arXiv:2206.14858, 2022.

[13] OpenAI, "GPT-4 Technical Report," arXiv preprint arXiv:2303.08774, 2023.

[14] J. Choi et al., "ChatGPT Goes to Law School," SSRN Electronic Journal, 2023. doi: 10.2139/ssrn.4335905.

[15] J. Choi et al., "A Cautionary Note on the Bar Exam Performance of GPT-4," SSRN Electronic Journal, 2023. doi: 10.2139/ssrn.4394099.

[16] A. Hope, M. Resnick, and D. Hu, "Assessing ChatGPT for Engineering Statics Analysis," arXiv preprint arXiv:2401.06720, 2024.

[17] T. Bommarito II et al., "GPT-4 and the USMLE: A Study of Clinical Knowledge and Reasoning," PLOS Digital Health, vol. 2, no. 2, e0000198, 2023.

[18] E. Kung et al., "Performance of ChatGPT on USMLE: Potential for AI-Assisted Medical Education," PLOS Digital Health, vol. 2, no. 2, e0000198, 2023.

[19] R. Li et al., "Competition-Level Code Generation with AlphaCode," Science, vol. 378, no. 6624, pp. 1092–1097, 2022.

[20] F. P. Beer, E. R. Johnston, Jr., D. F. Mazurek, P. J. Cornwell, and B. P. Self, *Vector Mechanics for Engineers: Statics and Dynamics*, 12th ed. New York, NY: McGraw-Hill Education, 2019.